\documentclass{article}

\usepackage{subcaption}

\usepackage{microtype}
\usepackage{graphicx}
\usepackage{booktabs} 

\usepackage{hyperref}

\usepackage[normalem]{ulem}
\usepackage{amsmath}
\usepackage{amssymb}
\usepackage{mathtools}

\DeclarePairedDelimiterX{\infdivx}[2]{(}{)}{%
  #1\;\delimsize\|\;#2%
}

\usepackage{bm}

\usepackage{multirow}

\usepackage{pgf}
\usepackage{tikz}
\usetikzlibrary{shapes,arrows,automata,backgrounds,calc}

\usepackage{array}
\newcolumntype{x}[1]{>{\centering\arraybackslash\hspace{0pt}}p{#1}}

\usepackage{colortbl}

\usepackage{hyperref}

\newtheorem{theorem}{Proposition}

\usepackage{balance}

\usepackage[accepted]{icml2019}

\icmltitlerunning{Trajectory-Based Off-Policy Deep Reinforcement Learning}

\begin{document}

\twocolumn[

\icmltitle{Trajectory-Based Off-Policy Deep Reinforcement Learning}

\begin{icmlauthorlist}
\icmlauthor{Andreas Doerr}{bosch,amd,mlr}
\icmlauthor{Michael Volpp}{bosch}
\icmlauthor{Marc Toussaint}{mlr}
\icmlauthor{Sebastian Trimpe}{amd}

\icmlauthor{Christian Daniel}{bosch}
\end{icmlauthorlist}

\icmlaffiliation{bosch}{Bosch Center for Artificial Intelligence, Renningen, Germany.}
\icmlaffiliation{amd}{Max Planck Institute for Intelligent Systems, Stuttgart/T{\"u}bingen, Germany.}
\icmlaffiliation{mlr}{Machine Learning and Robotics Lab, University of Stuttgart, Germany}

\icmlcorrespondingauthor{Andreas Doerr}{andreasdoerr@gmx.net}

\icmlkeywords{reinforcement learning, policy gradient, exploration, off-policy}

\vskip 0.3in
]

\printAffiliationsAndNotice{}

\begin{abstract}
	Policy gradient methods are powerful reinforcement learning algorithms and have been demonstrated to solve many complex tasks.
However, these methods are also data-inefficient, afflicted with high variance gradient estimates, and  frequently get stuck in local optima.
This work addresses these weaknesses by combining recent improvements in the reuse of off-policy data and exploration in parameter space with deterministic behavioral policies. 
The resulting objective is amenable to standard neural network optimization strategies like stochastic gradient descent or stochastic gradient Hamiltonian Monte Carlo. 
Incorporation of previous rollouts via importance sampling greatly improves data-efficiency, whilst stochastic optimization schemes facilitate the escape from local optima.
We evaluate the proposed approach on a series of continuous control benchmark tasks. 
The results show that the proposed algorithm is able to successfully and reliably learn solutions using fewer system interactions than standard policy gradient methods. 


	\vspace{-5mm}
\end{abstract}

\section{Introduction}
\label{sec:Introduction}

Policy search methods are amongst the few successful Reinforcement Learning (RL) \cite{sutton2000policy} methods which are applicable to high-dimensional or continuous control problems, such as the ones typically encountered in robotics \cite{peters2008reinforcement, deisenroth2013survey}.
One particular class of policy search methods directly estimates the gradient of the expected return with respect to the parameters of a differentiable policy. 
These Policy Gradient (PG) algorithms have achieved impressive results on highly complex tasks \cite{schulman2015trust, schulman2017proximal}.
However, standard algorithms are vastly data-inefficient and rely on millions of data points to achieve the aforementioned results.
Typical applications are therefore limited to simulated problems where policy rollouts can be cheaply obtained.

Algorithms based on stochastic policy gradients, like REINFORCE \cite{williams1992simple} and G(PO)MDP \cite{baxter2001infinite}, typically estimate the policy gradient based on a batch of trajectories, which are obtained by executing the current policy on the system (i.e. based on on-policy samples).
In the next step, all previous experience is discarded and new trajectories are sampled using the updated policy.
This scheme holds true also for more recent methods, like PPO \cite{schulman2017proximal} or POIS \cite{metelli2018policy}, where a surrogate objective is constructed, which can be optimized till convergence.
Typically, Importance Sampling (IS) techniques are employed to evaluate a target policy based on rollouts obtained from behavioural policies (i.e. from off-policy samples).
Albeit these off-policy evaluation schemes, in these algorithms, no data is shared between iterations.
Prominent examples of off-policy offline algorithms typically employ actor-critic architectures \cite{silver2014deterministic}, where the parametric critic model, typically a value function, is updated to summarize all knowledge gathered so far.
In contrast, we proposed the model-free Deep Deterministic Off-Policy Gradient method (DD-OPG)\footnote{\url{https://github.com/boschresearch/DD_OPG}}, which incorporates previously gathered rollout data by sampling from a trajectory replay buffer.
This effectively enables backtracking to promising solutions, whilst requiring only minimal assumptions to construct the surrogate model.

Next to the inefficient use of available data, stochasticity in both the policy and the environment causes highly variable gradient estimates and therefore slow convergence.
When executing the probabilistic policy on the system, noise is injected into the policy gradient in each time step, leading to a variance, which linearly increases with the length of the horizon \cite{munos2006policy}.
Additive Gaussian noise is typically employed as source of exploration.
Additionally, PG methods built around the likelihood ratio trick intrinsically require probabilistic policies.
Only then, policies can be updated to increase the likelihood of actions, which have been advantageous in previous rollouts.
Instead of independent noise, temporally-correlated noise \cite{osband2016deep}, or exploration directly in parameter space can lead to a larger variety of behaviours \cite{plappert2017parameter}.
Here, the behavioural policy is deterministic, thereby effectively reducing the gradient variance.
Methods like DPG \cite{silver2014deterministic} and DDPG \cite{lillicrap2015continuous} learn a parametric value function model to translate changes in policy and therefore actions to changes in expected value.
Similarly, our proposed model-free DD-OPG algorithm constructs a non-parametric critic based on importance sampling.
This critic, called surrogate model in the following, allows for updating a deterministic policy without the need for explicit parametric value models.

To summarize: We propose an importance sampling based surrogate model of the return distribution, which enables off-policy, offline policy optimization.
This surrogate facilitates deterministic policy gradients to reduce gradient variance and enables incorporation of all available data from a replay buffer.
Exploration in the policy parameter space is achieved by a prioritized resampling of the surrogates support data, thus favouring promising regions in policy space.
Normalized IS, which we demonstrate to act similarly as a baseline in standard PG methods, additionally reduces the variance of the employed estimates.
Although no additional, parametric value function baseline (as utilized in TRPO/PPO for variance reduction) is required in our method, fast progress and therefore data-efficient learning is demonstrated on typical continuous control tasks.

The general problem formulation and policy gradient framework is highlighted in Sec.\,\ref{sec:Preliminaries}, followed by a short presentation of the standard importance sampling estimators to incorporate off-policy data in Sec.\,\ref{sec:OffPolicyEvaluation}.
The surrogate model, necessary to efficiently incorporate deterministic policy data, as the core of the proposed model-free DD-OPG method is detailed in Sec.\,\ref{sec:DeterministicPolicyGradients}.
In Sec.\,\ref{sec:ModelFreeOffPolicyOptimization}, the main policy optimization scheme is presented and experimentally evaluated in Sec.\,\ref{sec:ExperimentalEvaluation}.
This work closes with a discussion of connections to related work in Sec.\,\ref{sec:RelatedWork} and concludes with an outlook into future work and open topics in Sec.\,\ref{sec:Discussion}.
\section{Preliminaries}
\label{sec:Preliminaries}

This section depicts the general episodic RL problem in a discrete-time Markovian environment and summarizes as core building-block of the proposed DD-OPG method, the standard return based policy gradient estimators \cite{williams1992simple}.
DD-OPG closely follows this algorithmic structure (cf. Alg.\,\ref{alg:DDOPG}), however with extensions to incorporate deterministic, off-policy rollouts as detailed in the following sections.
The RL problem is characterized by a discrete-time Markov Decision Process (MDP) $\mathcal{M} = (\mathcal{S}, \mathcal{A}, p, r, \gamma, p_0)$.
An agent is interacting with an environment, whose states $s_t \in \mathcal{S}$ transitions according to the agent's actions $a_t \in \mathcal{A}$ and the environment's transition probabilities $p(s_{t+1}\mid s_t, a_t)$ into a successor state.
Starting from a state $s_0$ drawn from the initial state distribution $p(s_0)$, agent tries to maximize its discounted reward, according to a reward function $r: S \times A \rightarrow \mathbb{R}$ and discount factor $\gamma$, accumulated over a horizon length $H$.
In policy search, the agent acts according to a (stochastic) policy $\pi_\theta = \pi(a_t \mid s_t; \theta)$, parameterized by $\theta$.
The expected accumulated reward is given by
\begin{equation}
J(\theta) = \int p(\tau \! \mid \! \theta) R(\tau) d\tau\,,
\label{eq:ExpectedCostDefinition}
\end{equation}
where the trajectory $\tau \in \mathcal{T}$ is the sequence of state-action pairs $\tau = (s_0, a_0, \ldots, s_H, a_H)$, the (discounted) trajectory return is given by $R(\tau) = \sum_{t=0}^{H-1} \gamma^t r(s_{\tau, t}, a_{\tau, t})$, and due to the Markov property, the trajectory distribution in \eqref{eq:ExpectedCostDefinition} is given by
\begin{equation}
p(\tau \! \mid \! \theta) = p(s_0) \prod_{t=0}^H p(s_{t+1} \! \mid \! s_t, a_t) \pi(a_t \! \mid \! s_t; \theta)\,.
\label{eq:TrajectoryLikelihood}
\end{equation}
The dynamics of the system $p(s_{t+1} \mid s_t, a_t)$ and the initial state distribution $p(s_0)$ are generally unknown to the learning agent.

Model-free policy gradient methods typically directly estimate the expected cost gradient based on the log-derivative trick.
The gradient is given by
\begin{equation}
\nabla_\theta J(\theta) = \int p(\tau \! \mid \! \theta) \nabla_\theta \log p(\tau \! \mid \! \theta) R(\tau) d\tau\,.
\label{eq:LikelihoodRatioPolicyGradient}
\end{equation}
Given on-policy samples $\tau_i \sim p(\tau | \theta)$, the following Monte Carlo (MC) estimators are obtained for the expected return
\begin{equation}
\hat{J}^{MC} (\theta) = \frac{1}{N} \sum_{i=1}^N R(\tau_i)\,,
\label{eq:J_MC}
\end{equation}
and the policy gradient
\begin{equation}
\hat{\nabla_\theta J^{MC}}(\theta)\! =\! \frac{1}{N} \sum_{i=1}^N \! \left[\sum_{t=0}^H \nabla_\theta \log \pi(a_t \! \mid \! s_t; \theta) R(\tau_i)\right]\!.
\label{eq:J_grad_MC}
\end{equation}
Since the unknown initial state and dynamics distributions are independent of the policy parameters $\theta$ (cf.\ \eqref{eq:TrajectoryLikelihood}), the trajectory likelihood gradient $\nabla_\theta \log p(\tau \! \mid \! \theta)$ with respect to the policy parameters can be computed analytically for a given, differentiable policy $\nabla_\theta \log \pi(a_t \! \mid \! s_t; \theta)$.
\section{Off-Policy Evaluation}
\label{sec:OffPolicyEvaluation}

The MC estimators require a substantial amount of on-policy rollouts $\tau_i \sim p(\tau|\theta^*)$ to reduce the gradient estimator's variance and typically many more rollouts than used in state-of-the-art implementations to closely approximate the true gradient \cite{ilyas2018deep}.

For off-policy data, Importance Sampling (IS) can be utilized to incorporate trajectories from a behavioural policy $\pi_{\theta'}$ in order to evaluate a new target policy $\pi_{\theta*}$ \cite{zhao2013efficient, espeholt2018impala, munos2016safe, metelli2018policy}.
In general, a Monte Carlo estimate of an expectation $\int p(x) f(x) dx$ (such as \eqref{eq:ExpectedCostDefinition}) can be obtained by sampling from a tractable distribution $x_i \sim q(x)$ and re-weighting the sampled function evaluations $f(x_i)$ based on the likelihood-ratio $p(x_i) / q(x_i)$.
The expected return can be rewritten as
\begin{equation}
J(\theta) = \int p(\tau \mid \theta') \frac{p(\tau \mid \theta)}{p(\tau \mid \theta')} R(\tau) d\tau\,,
\label{eq:ImportanceSamplingGeneral}
\end{equation}
such that the IS weighted Monte Carlo estimator is given by
\begin{align}
\hat{J}^{IS}(\theta) &= \frac{1}{N} \sum_{i=0}^N \frac{p(\tau_i \mid \theta)}{p(\tau_i \mid \theta')} R(\tau_i) \\
&= \frac{1}{N} \sum_{i=0}^N w(\tau_i, \theta) R(\tau_i)\,,
\label{eq:MCISCostEstimator}
\end{align}
where $N$ trajectories are sampled from a policy $\pi_{\theta'}$ to infer the expected cost of policy $\pi_{\theta}$.
Although system dynamics and initial state distribution in \eqref{eq:TrajectoryLikelihood} are unknown, the likelihood-ratio, i.e. the importance weights, can be computed since the unknown parts cancel out, such that
\begin{equation}
w(\tau, \theta) = \frac{p(\tau \mid \theta)}{p(\tau \mid \theta')} = \frac{\prod_{t=0}^H \pi(a_t \mid s_t; \theta)}{\prod_{t=0}^H \pi(a_t \mid s_t; \theta')}.
\label{eq:ImportanceWeights}
\end{equation}

During learning, trajectories are collected from multiple different policies $\mathcal{D} = \{(\tau_i, \theta_i)\}_{i=1}^N$.
To incorporate all data, the importance sampling distribution can be replaced by an empirical mixture distribution $q(\tau \mid \theta_1, \ldots, \theta_N) = 1 / N \sum_i p(\tau \mid \theta_i)$ such that the available trajectories are i.i.d. draws from the empirical mixture distribution $\tau_i \sim q(\tau \mid \theta_1, \ldots, \theta_N)$ \cite{jie2010connection}.
The resulting importance weights are given by
\begin{equation}
w(\tau, \theta) = \frac{\prod_{t=0}^H \pi(a_t \! \mid \! s_t; \theta)}{\frac{1}{N}\sum_j \prod_{t=0}^H \pi(a_t \! \mid \! s_t; \theta_j)}\,.
\label{eq:MixtureImportanceWeights}
\end{equation}
Computing the importance weights in \eqref{eq:MixtureImportanceWeights}, however, scales quadratically with the number of available trajectories due to the summation over the likelihoods of all trajectories given all available policies.
Scaling this estimator to today's deep neural network policies with a large number of required rollouts is, thus, a major challenge.
Instead of computing the surrogate based on all data, as in \cite{jie2010connection}, which is only feasible for several hundred rollouts, the proposed DD-OPG method employs a trajectory replay buffer and a probabilistic selection scheme to recompute a stochastic approximation of the full surrogate model.
This idea is related to prioritized experience replay \cite{schaul2015prioritized} but for full trajectories.
It enables scaling to much larger datasets and at the same time helps to avoid local minima by stochastically optimizing the objective.

Another technique typically employed for IS is weight normalization \cite{metelli2018policy}.
The \emph{weighted importance sampling} estimator obtains a lower variance estimate at the cost of adding bias.
It has been employed in \cite{peshkin2002learning} and is both theoretically and empirically better-behaved \cite{meuleau2000off, precup2000eligibility, shelton2001policy} compared to the pure IS estimator.
The weighted importance sampling estimator is given by
\begin{equation}
\hat{J}^{\text{WIS}}(\theta) = \frac{1}{Z} \sum_{i=0}^N w(\tau_i, \theta) R(\tau_i)\,,
\label{eq:WeightedImportanceSampling}
\end{equation}
where importance weights $w(\tau_i, \theta)$ might be computed according to \eqref{eq:ImportanceWeights} or \eqref{eq:MixtureImportanceWeights} and a normalizing constant $Z = \sum_{i=0}^N w(\tau_i, \theta)$ instead of the standard normalization $Z = N$, previously used in \eqref{eq:MCISCostEstimator}.

From the policy gradient perspective, by normalizing the importance weights, we obtain a gradient estimator, which includes a parameter dependent baseline.
\begin{theorem}
\label{th:NormalizedPolicyGradient}
The policy gradient estimator obtained from the self-normalized importance sampling expected cost estimator $\hat{J}^{WIS}$ is given by
\begin{align}
    \label{eq:NormalizedPolicyGradient}
    \nabla_\theta \hat{J}^{\text{WIS}}&(\theta) \! = \frac{1}{Z} \sum_{i=1}^N \bigg[ w(\tau_i, \theta) \\ 
    & \sum_{t=0}^H \left[ \nabla_\theta \log \pi(a_t^{(i)}  \! \mid \! s_t^{(i)}; \theta) \right] \! 
    \left[ R(\tau_i) \! -\!  \hat{J}^{\text{WIS}}(\theta) \right] \bigg]\,. \nonumber
\end{align}
\end{theorem}
A proof of this proposition is shown in Appendix A.
This estimator is closely related to standard PG estimators with an added baseline term for variance reduction.

In standard, REINFORCE like, PG methods, two of the most common variance reduction techniques \cite{greensmith2004variance} are:
i) incorporation of the reward-to-go for each policy action update instead of the entire Monte Carlo path return; and
ii) subtraction of a state dependent baseline term, such as to obtain an estimate of the advantage of the previously taken action.
The intuition behind method i) is to reward actions only for rewards obtained after the action took effect, but not for those obtained earlier on.
However, to compute the importance weights not for the full trajectory distribution but for each state-action pair individually, the computation of a matrix of size $\mathcal{O}(N^2H^2)$ would be required.
Therefore, the model-free, importance sampling based approaches are typically limited to the path return based estimators.
Model-based methods (i.e. a parametric models for the value function) are employed in the cost-to-go estimators.
Variance reduction method ii) is automatically obtained by the normalized estimator as shown in proposition \ref{th:NormalizedPolicyGradient}, however, in contrast to the bias free value function control variates, at the cost of adding bias.
Additional, optimal baselines to further decrease the variance of the gradient estimator have been derived in \cite{jie2010connection} and could be incorporated into DD-OPG.
\section{Deterministic Policy Gradients}
\label{sec:DeterministicPolicyGradients}

The policy gradient estimators in \eqref{eq:J_grad_MC} and \eqref{eq:NormalizedPolicyGradient} rely on a policy distribution $\pi(a_t | s_t; \theta)$ in order to obtain a gradient signal on how to update the policy parameters to increase the likelihood of successful actions.
In this situation the, typically Gaussian, additive policy noise acts in two ways, causing exploration and serving as the basis for the estimation of the objective function.
\\ %
\emph{Exploration} is being driven directly through noise in the action space, i.e., the policy covariance. 
While driving exploration through noisy actions will converge in the limit, the resulting explorative behaviour exhibits no temporal correlations, which can make it inefficient. 
\\ %
\emph{Estimation} of the objective function is typically achieved by reweighting the action distribution according to the policy's likelihood.
Standard policies are given as 
$\pi(a_t | s_t; \theta) = \mathcal{N}(a_t | \mu_{\theta}(s_t), \Sigma_\theta)$, 
where $\mu_{\theta}$ is represented by some function approximator parameterized by $\theta$, e.g. a neural network.
The additive Gaussian noise covariance is typically a diagonal matrix, parameterized by $\theta$ as well.
The proposed deterministic policy gradient method strives to separate the exploration and estimation part.

\emph{Parameter Space Exploration} 
By utilizing deterministic rollout policies, the only noise introduced into the gradient estimate originates from the stochasticity of the environment and we have to perform exploration in parameter space instead of action space exploration. However, as stated above, parameter based exploration may in many cases be more efficient than exploration in action space, since parameter based exploration will lead to temporally correlated actions which can explore the state space faster. 
Typically, however, this effect is negated for neural network policies since the parameter space that has to be explored is prohibitively large. 
Thus, to navigate large parameter spaces efficiently, some approximate evaluation of the cost function \eqref{eq:ExpectedCostDefinition} is needed.


\emph{Trajectory based objective estimate}
Whilst evaluation of the Monte Carlo based expected cost estimate is possible also for deterministic policies, the off-policy evaluation is no longer feasible since the likelihood ratio $p(\tau | \theta) / p(\tau | \theta')$ (cf.\,\eqref{eq:ImportanceWeights}) becomes zero for two distinct dirac policy action distributions if $\mu_\theta(s) \neq \mu_{\theta'}(s)$.

However, we can still compare trajectories under a stochastic evaluation distribution, similar to a kernel  function where the standard deviation of the evaluation function relates to a kernel lengthscale in action space. 

Thus, we introduce the evaluation policy 
\begin{equation}
    \tilde{p}(a_t | s_t; \theta) = \mathcal{N}(a_t | \mu_\theta (s_t), \Sigma)\,,
    \label{eq:EvaluationPolicy}
\end{equation}
where $\Sigma = \text{diag}(\sigma_1, \ldots, \sigma_{D_u})$ is a diagonal covariance matrix as typically employed in deep RL methods with Gaussian action noise.
The deterministic policy is given by $p(a_t | s_t; \theta) = \delta(a = \mu_\theta(s_t))$, where $\delta$ is the dirac delta.
From the general IS expectation in \eqref{eq:ImportanceSamplingGeneral} and our evaluation policy in \eqref{eq:EvaluationPolicy}, the surrogate model follows as
\begin{equation}
\hat{J}^{\text{surr}}(\theta) =
\frac{1}{Z}
\sum_{i=1}^N \tilde{w}(\tau_i, \theta) R(\tau_i) \,,
\label{eq:SurrogateModel}
\end{equation}
with surrogate weights
\begin{equation}
\tilde{w}(\tau_i, \theta) = \frac{\prod_{t=0}^H \mathcal{N}(a^{(i)}_t | \mu_{\theta}(s^{(i)}_t), \Sigma)}{\frac{1}{N} \sum_{j=0}^N \prod_{t=0}^H \mathcal{N}(a^{(i)}_t | \mu_{\theta_j}(s^{(i)}_t), \Sigma)}, 
\label{eq:SurrogateWeights}    
\end{equation}
where, depending on the choice of normalization constant $Z$, we obtain the analogue to the standard IS estimator ($Z\! =\! N$) or the analog to the weighted IS estimator ($Z \!=\! \sum_{i=1}^N \tilde{w}(\tau_i, \theta)$).
Reintroducing the fixed Gaussian noise as an implicit loss to obtain gradients for the evaluation of deterministic policies is clearly a model assumption in the proposed method but can be justified from several perspectives.

The hyper-parameter $\Sigma$ allows for control over the amount of information shared between neighbouring policies.
Similar to the cap of importance weights in PPO \cite{schulman2017proximal}, this parameter allows to control bias and variance of the surrogate model.
Analyzing the introduced bias and relation to the PPO weight cap is however ongoing research.
In the limit of $\Sigma \rightarrow \mathbf{0}$, the proposed surrogate \eqref{eq:SurrogateModel} approaches the MC estimator \eqref{eq:J_MC}.
Only in case of two different policy parameterizations $\theta_j \neq \theta_i$, but equivalent actions $\mu_{\theta_j}(s) = \mu_{\theta_i}(s)$ for the sampled states $s$, the surrogate model would output an average whereas the MC estimator would not mix up the obtained returns.
For $\Sigma = \Sigma_\theta$, the surrogate model recovers the true IS estimate, given that all trajectories are generated using the same additive Gaussian noise.
Finally, for $\Sigma \rightarrow \inf$, the estimate is simply the average over all available path returns.

Modelling the expected return distribution by choosing a lengthscale in action space can furthermore be motivated from a second perspective.
Typical expected return distributions oftentimes comprise sharp transitions between stable and unstable regions, where policy parameters change only slightly but reward changes drastically.
One global lengthscale is therefore typically not well suited to directly model the expected return.
This is a standard problem in Bayesian Optimization for reinforcement learning, where typical smooth kernel functions (e.g. squared exponential kernel) with globally fixed lengthscales are unable to model both stable and unstable regimes at the same time.
However, in the proposed model, a lengthscale in action space is translated via the sampled state distribution and policy function $\mu$ into implicit assumptions in the actual policy parameter space.
Doing so, instead of operating on arbitrary euclidean distances in policy parameter space, a more meaningful distance in trajectory and action space is available.
Typically, for a given system, distance of trajectories and between actions is more graspable, compared to arbitrary deep neural network policy parameters.


The expected return estimator \eqref{eq:SurrogateModel} falls back to zero for policy evaluation far away from training data.
To estimate the variance of the importance sampling estimator itself, typically, the Effective Sample Size (ESS) is evaluated.
Based on the variance of the importance weights, it analyses the effective number of available data points at a specific policy evaluation position.
In \cite{metelli2018policy}, a lower bound on the expected return has been proposed such that with probability $1-\delta$ it holds that
\begin{align}
    \text{E}_{\tau \sim p(\tau | \theta)}[R(\tau)] \geq & \frac{1}{N}\sum_{i=1}^N \tilde{w}(\tau_i, \theta) R(\tau_i) \nonumber \\
    & - \|R\|_\infty \sqrt{\frac{(1 - \delta) d_2 (p(\tau | \theta) \| p(\tau | \theta'))}{\delta N}}\,,
    \label{eq:LowerBound}
\end{align}
where $d_2$ is the exponentiated 2-R\'{e}nyi divergence.
Due to the identity $\text{ESS}(P || Q) = N / d_2(P || Q)$, this lower bound can be estimated in a sample-based way by employing the ESS estimator
\begin{equation}
    \hat{\text{ESS}} = \frac{1}{\sum_{i=1}^N \tilde{w}(\tau_i, \theta)^2}\,,
\end{equation}
such as to obtain the lower bound estimate
\begin{align}
    \text{E}_{\tau \sim p(\tau | \theta)}[R(\tau)] \geq 
    &\frac{1}{N}\sum_{i=1}^N \tilde{w}(\tau_i, \theta) R(\tau_i) \\
    & - \|R\|_\infty \sqrt{\frac{1 - \delta}{\delta} \text{ESS}(\theta)^{-1}}\,.
    \label{eq:ESSLowerBound}
\end{align}
Refer to theorem 4.1 in \cite{metelli2018policy} for details and proof regarding the lower bound in \eqref{eq:LowerBound}.
The confidence parameter $\delta$ determines, similar to the KL-divergence in TRPO \cite{schulman2015trust}, how far the policy optimization can step away from known regions.
In DD-OPG, this uncertainty estimate is employed as penalty
\begin{equation}
    \text{penalty}(\theta) = - \| R \|_\infty \gamma \sqrt{\hat{\text{ESS}}(\theta)^{-1}}\,,
\end{equation}
with penalty factor $\gamma$ as an hyper-parameter to control exploration, i.e. following the objective estimate vs. risk awareness, i.e. staying within a trust region.
\section{Model-Free Off-Policy Optimization}
\label{sec:ModelFreeOffPolicyOptimization}

\begin{figure*}
    \centering
    \begin{subfigure}{0.32\textwidth}
        \includegraphics{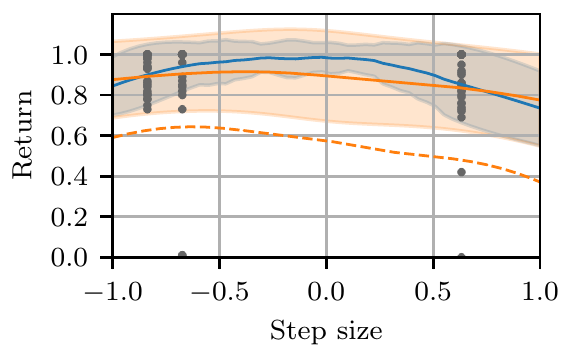}
        \caption{$\log(\Sigma) = 0 \cdot \mathbf{I}$}
        \label{fig:SurrogatePrediction0}
    \end{subfigure}
    \hfill
    \begin{subfigure}{0.32\textwidth}
        \includegraphics{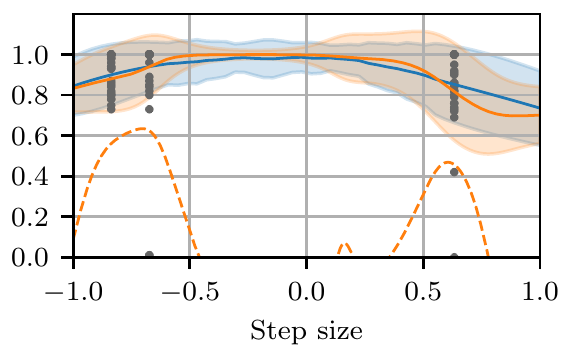}
        \caption{$\log(\Sigma) = -1 \cdot \mathbf{I}$}
        \label{fig:SurrogatePrediction-1}
    \end{subfigure}
    \hfill
    \begin{subfigure}{0.32\textwidth}
        \includegraphics{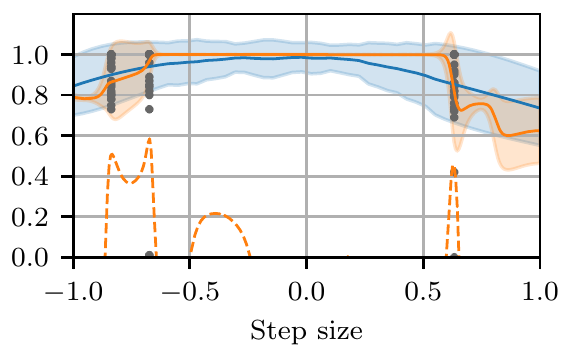}
        \caption{$\log(\Sigma) = -2 \cdot \mathbf{I}$}
        \label{fig:SurrogatePrediction-2}
    \end{subfigure}
    \caption{Visualization of the surrogate return model. A cross-section along a random direction in parameter space is shown for parameters close to optimum. The ground truth mean and std (blue) of the return distribution is shown together with the mean and std estimate (orange) from the weighted importance sampling surrogate model. The lower confidence bound ($\delta = 0.2$, dashed orange line) is shown together with the model's input data (grey dots). Notice how more or less information is shared between points where data is available depending on the chosen lengthscale parameter $\Sigma$.}
    \label{fig:SurrogatePrediction}
\end{figure*}

The surrogate model of the return distribution, as derived in Sec.\,\ref{sec:DeterministicPolicyGradients}, can now be directly incorporated for policy optimization.
In related work, parametric search distributions (e.g. Gaussian) are employed as policy search distribution or hyperpolicy \cite{zhao2013efficient, plappert2017parameter, metelli2018policy}.
However, in high-dimensional spaces, as typically obtained with deep network policy representations, updating the full search distribution is challenging and common approaches usually revert to heuristics to control a simplified, e.g. diagonal or block-wise search distribution's covariance matrix.

Instead, the proposed model-free DD-OPG method fully optimizes a stochastic version of the surrogate objective to foster exploration and overcome local minima.
At the same time, the stochastic evaluation mitigates the unfavourable complexity of computing the full importance sampling estimate based on all available data.
Due to the empirical mixture distribution in \eqref{eq:MixtureImportanceWeights}, computing the likelihood of all observed trajectories under all policies is quadratic in the number of observed paths.
Instead, the proposed method employs a selection criterion to construct a stochastic surrogate model based on a subset of rollouts in each policy optimization step.
In particular, a predefined number of $N_{max}$ rollout indices is drawn from the softmax distribution over the discrete set of available trajctory indices $\mathcal{I}$.
The softmax is computed based on the normalized, empirical returns $\tilde{R}$ and a temperature factor $\lambda$.
\begin{equation}
    p(\mathcal{I} | \tau_1, \ldots, \tau_N) = \frac{\exp (\tilde{R}(\tau_\mathcal{I}) / \lambda)}{\sum_{j=1}^N \exp (\tilde{R}(\tau_j) / \lambda)}.
\end{equation}
The temperature $\lambda$ is used to trade off exploration against exploitation in the selection of reference trajectories.
This scheme is closely related to prioritized experience replay \cite{schaul2015prioritized}.
A study of the effect of temperature selection on the learning progress is shown in Sec.\,\ref{sec:AblationStudy}.

The full DD-OPG algorithm is detailed in Alg.\,\ref{alg:DDOPG}.
The main objective is to incorporate all available deterministic policy rollouts, not only the ones from the current iteration, into the surrogate model by means of the softmax replay selection.
The lower bound expected return can then be fully optimized using standard optimization techniques.
In practice Adam \cite{kingma2014adam} is employed, but other techniques, e.g. based on the natural policy gradient \cite{peters2008natural} could be incorporated as well.

\begin{algorithm}[tb]
   \caption{Model-free DD-OPG}
   \label{alg:DDOPG}
\begin{algorithmic}
   \STATE {\bfseries Input:} Initial policy parameters $\theta_0$
   \STATE Empty trajectory replay buffer $\mathcal{D}_0 = \{\}$
   \REPEAT 
       \STATE Sample trajectory: $\tau_i \sim p(\tau \! \mid \! \theta_i)$
       \STATE Update trajectory buffer: $\mathcal{D}_{i+1} = \mathcal{D}_i \cup (\tau_i, R_i)$
       \STATE Memory selection: $i_1, \ldots, i_{N_{max}} \overset{iid}{\sim} p(\mathcal{I} \! | \! \tau_1, \ldots, \tau_i)$
       \STATE Surrogate model: $\tilde{J}(\theta)$, $\text{penalty}(\theta)$
       \STATE Lower bound optimization:
       \STATE $\theta_{i+1} = \underset{\theta}{\text{argmax}} \tilde{J}(\theta) - \text{penalty}(\theta)$
   \UNTIL{converged or maximum iterations}
\end{algorithmic}
\end{algorithm}
\section{Experimental Evaluation}
\label{sec:ExperimentalEvaluation}

The experimental evaluation of the proposed DD-OPG method is threefold.
In Sec.\,\ref{sec:SurrogateModel}, the resulting surrogate return model is visualized, highlighting different modeling options.
A benchmark against state-of-the-art PG methods is shown in Sec.\,\ref{sec:PolicyGradientBenchmark} to highlight fast and data-efficient learning.
Finally, important parts of the proposed algorithms and their effects on the final learning performance are highlighted in an ablation study in Sec.\,\ref{sec:AblationStudy}.

\subsection{Surrogate Model}
\label{sec:SurrogateModel}

As discussed in Sec.\,\ref{sec:DeterministicPolicyGradients}, the proposed surrogate model can smoothly interpolate between the Monte Carlo estimate, the importance sampling estimate, and an average of all available returns.
In Fig.\,\ref{fig:SurrogatePrediction}, the available surrogate model predictions are visualized for multiple settings of the model hyper-parameter $\Sigma$.
In particular, the estimate for expected return (solid orange line), return variance (shaded orange visualizes one standard deviation), and the lower bound of the expected return (dashed orange line) are visualized for policy evaluations along a random direction around the optimal policy $\theta^*$ for the cartpole environment (experimental details can be found in Appendix B).
Trajectory data, which is available to the estimator is highlighted by grey dots.
The groundtruth return distribution (mean +/- one std. in blue) is computed using the standard MC estimator, based on independent policy rollouts, which are \emph{not} part of the surrogate model.

Stepping from long lengthscales (cf.\,Fig.\,\ref{fig:SurrogatePrediction0}) to shorter lengthscales (cf.\,Fig.\,\ref{fig:SurrogatePrediction-2}), the surrogate model predictions become more local.
Most visibly in the lowerbound estimate, the ESS drops significantly when moving away from data points and small model lengthscales, resulting in much higher uncertainty.

\subsection{Policy Gradient Benchmark}
\label{sec:PolicyGradientBenchmark}

\begin{figure*}
    \centering
    \begin{subfigure}{0.32\textwidth}
        \includegraphics{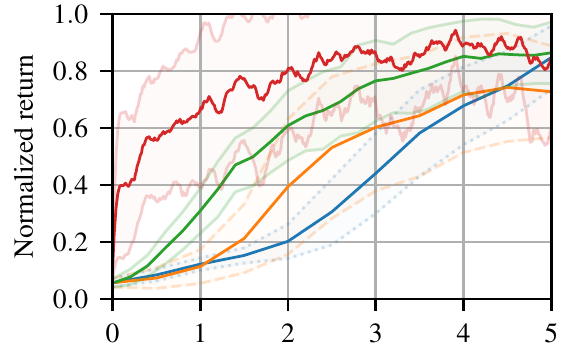}
        \caption{Cartpole, [-] $\times 10^5$ steps}
        \label{fig:BenchmarkCartpole}
    \end{subfigure}
    \hfill
    \begin{subfigure}{0.32\textwidth}
        \includegraphics{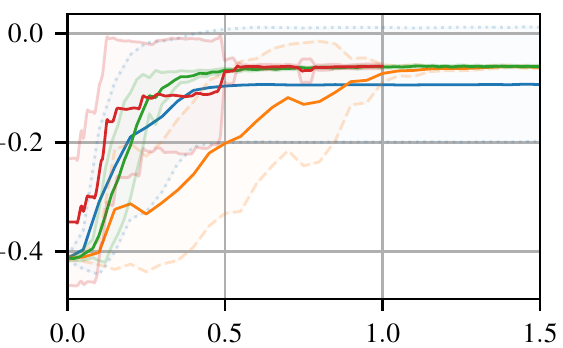}
        \caption{Mountaincar, [-] $\times 10^5$ steps}
        \label{fig:BenchmarkMountaincar}
    \end{subfigure}
    \hfill
    \begin{subfigure}{0.32\textwidth}
        \includegraphics{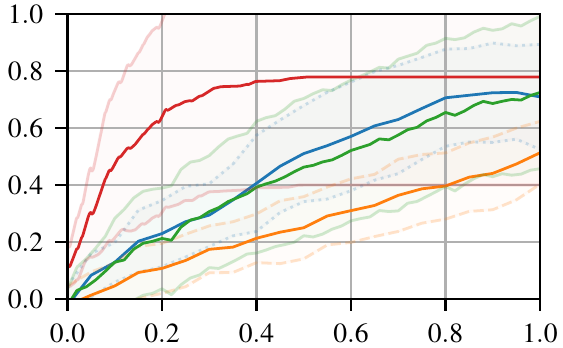}
        \caption{Swimmer, [-] $\times 10^5$ steps}
        \label{fig:BenchmarkSwimmer}
    \end{subfigure}
    \caption{Policy gradient methods benchmark. The proposed method DD-OPG (red) is compared to standard REINFORCE (blue), TRPO (orange) and PPO (green) on three continuous control benchmark problems. Mean and standard deviation of the average return (obtained from 10 independent random seeds) are plotted as a function of the system interaction steps (scaled by $10^5$). Significant faster learning speed in the beginning is observed for the model-free off-policy method in comparison to the on-policy PG methods.}
    \label{fig:Benchmark}
    \vskip -0.2in
\end{figure*}

The proposed DD-OPG method is evaluated in terms of data-efficiency and learning progress in comparison to state-of-the-art policy gradient methods based on Monte Carlo return estimates.
In contrast, methods such as DDPG \cite{lillicrap2015continuous} employ TD learning for their value function model and are not part of this evaluation.
The benchmark compares DD-OPG to the standard REINFORCE \cite{williams1992simple} baseline and both TRPO \cite{schulman2015trust} and PPO \cite{schulman2017proximal}.
All competitor algorithms employ, as it is common practice, the reward-to-go formulation and a linear feature-based baseline for variance reduction.
For all methods, hyper-parameters are selected to achieve maximal accumulated average return, i.e. fast and stable policy optimization.
Details about the individual methods' configuration and the employed environments can be found in Appendix B.

The resulting learning performances are visualized in Fig.\,\ref{fig:Benchmark} for the cartpole, mountaincar and swimmer environment (left to right) \cite{duan2016benchmarking}.
For REINFORCE (blue), TRPO (yellow), PPO (green), and DD-OPG (red), the mean average return (solid line) and its confidence intervals (one standard deviation as shaded area) are depicted, as obtained from 10 independent runs out of 10 random seeds for each environment and method.
To compare the learning speed and data-efficiency between the batch-wise learning competitors and the rollout-based DD-OPG, the results are visualized as a function of collected environment interactions (scaled by $10^5$) in Fig.\,\ref{fig:Benchmark}.

With DD-OPG, rapid learning progress is achieved already and the final performance of the competitive, state-of-the-art policy gradient methods is matched.
In the hyper-parameter tuning phase, experiments with TRPO and PPO have been conducted based on smaller batchsizes, but due to the lack of data-efficient incorporation of off-policy data, no faster \emph{and} stable learning progress could be achieved for these methods, compared to the one visualized in Fig.\,\ref{fig:Benchmark}.
Notice the large variance of the DD-OPG learning progress in the swimmer environment.
Albeit the superior learning performance of DD-OPG on the swimmer environment, some of the runs got stuck in local minima, resulting in the large variance estimate.
This trade-off between exploration and exploitation is partially achieved by the stochastic memory selection.
A mix of prioritized trajectory replay and current trajectories is mandatory to prevent greedy exploitation of previously seen, local minima and to facilitate exploration.
Our experiments show that it is mandatory to incorporate previously seen rollout data, as it is done in DD-OPG, to enable rapid progress already in the early stages of training.

\subsection{Ablation Study}
\label{sec:AblationStudy}

In the final DD-OPG algorithm, multiple aspects come together: i) the deterministic surrogate model, ii) the memory selection strategy, and iii) the optimization scheme.
In this ablation study, we separate the individual components to analyse their effect on the final learning performance.
Experiments are conducted on the cartpole environment and results are averaged over three random seeds.

\begin{figure}[ht]
\vskip 0.2in
\begin{center}
\centerline{\includegraphics[width=\columnwidth]{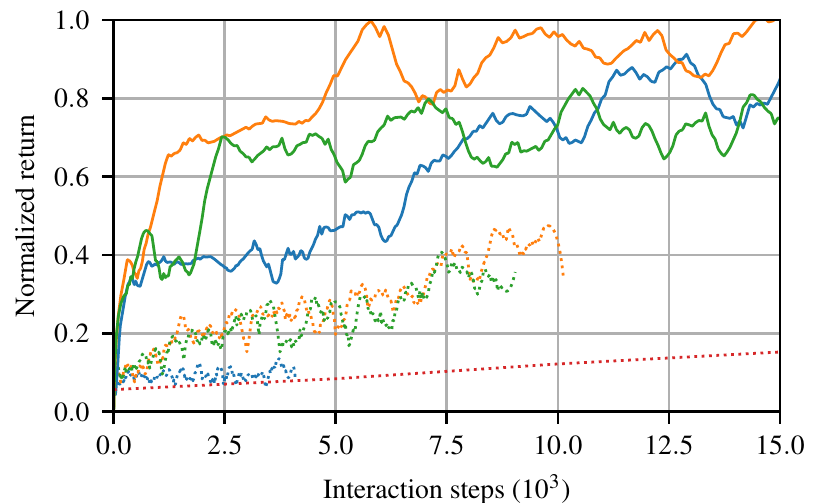}}
\caption{Ablation study of DD-OPG. The full DD-OPG model is constructed from the REINFORCE baseline by iteratively adding i) deterministic off-policy data incorporation and ii) full optimization of the surrogate model. Visualized is the mean learning progress from 3 random seeds on the cartpole environment. REINFORCE (dashed red line) is shown with DD-OPG optimizing only for one gradient step (dotted lines) and fully optimizing the surrogate model (solid lines). For DD-OPG, three levels of history are shown (blue: $N_{max} = 5$, green: $N_{max} = 20$, yellow: $N_{max} = 50$).} 
\label{fig:AblationHistory}
\end{center}
\vskip -0.4in
\end{figure}

\begin{figure}[ht]
\vskip 0.2in
\begin{center}
\centerline{\includegraphics[width=\columnwidth]{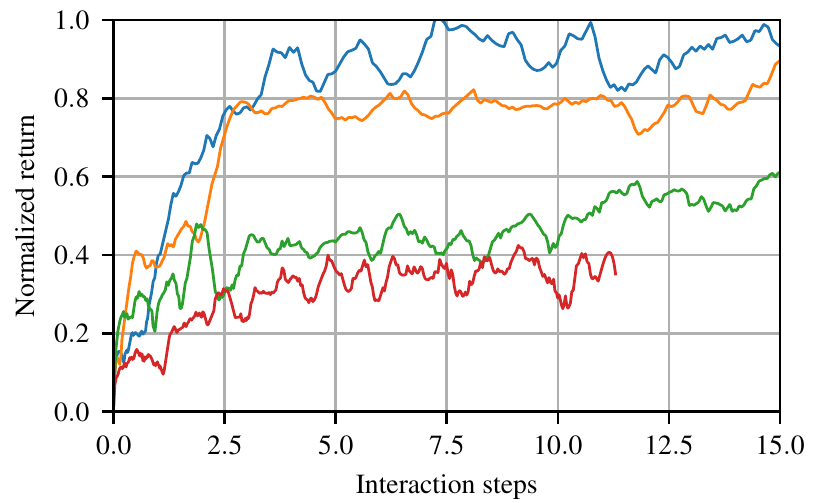}}
\caption{Effect of the surrogate hyper-parameter $\Sigma$ on the learning progress. Learning speed increases from short lengthscales $\log \Sigma = 1.0$ (red) to $\log \Sigma = 2.0$ (green), $\log \Sigma = 3.0$ (yellow), and $\log \Sigma = 4.0$ (blue). Visualized are DD-OPG mean learning curves from three random seeds as a function of the number of interaction steps with the cartpole environment (scaled by $10^3$).} 
\label{fig:AblationModel}
\end{center}
\vskip -0.3in
\end{figure}

\begin{figure}[ht]
\vskip 0.2in
\begin{center}
\centerline{\includegraphics[width=\columnwidth]{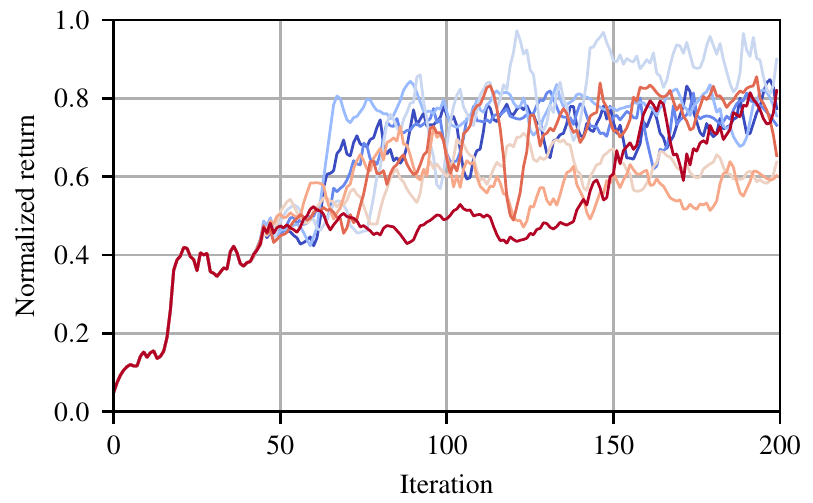}}
\caption{Learning progress for multiple temperature settings for softmax trajectory selection. From lowest temperature ($\lambda$ = 0.01, blue) to highest temperature ($\lambda$ = 2.0, red). Both too high and too low temperatures lead to suboptimal behaviour, either by too much exploration or too greedy behaviour.} 
\label{fig:AblationTemp}
\end{center}
\vskip -0.4in
\end{figure}

In the first experiment, DD-OPG is reconstructed starting from the REINFORCE baseline.
A visualization is shown in Fig.\,\ref{fig:AblationHistory}.
In REINFORCE (red dotted line), only one policy gradient step is taken based on the current on-policy data.
This is comparable to DD-OPG with almost no memory ($N_{max} = 5$) and only one step gradient update (visualized as blue dotted line).
Learning performance is already increased by adding more memory paths (green: $N_{max} = 20$, yellow: $N_{max} = 50$).
More significantly, the full optimization of the surrogate model (solid lines) achieves much faster learning progress.

In Fig.\,\ref{fig:AblationModel}, the effect of the surrogate model's lengthscale parameter $\Sigma$ is evaluated.
Four different lengthscales $\log \Sigma$ are evaluated (red: 1.0, green: 2.0, yellow: 3.0, blue: 4.0).
In this experiment, longer lengthscales clearly improve learning speed despite the introduced model bias.

The effects of the softmax temperature $\lambda$ on the proposed prioritized trajectory replay and the learning progress are depicted in Fig.\,\ref{fig:AblationTemp}. 
Explorative behaviour is favoured for higher temperatures (red), whereas for low temperatures (blue), previous trajectories are selected more greedily.
In this example, an intermediate temperature achieves the best trade-off exploration-exploitation trade-off. 
\section{Connections to Related Work}
\label{sec:RelatedWork}

Policy search methods \cite{peters2008reinforcement, deisenroth2013survey} and policy gradient methods \cite{williams1992simple, baxter2001infinite} are well studied in the RL community and many connections to DD-OPG exist.

Importance sampling has been employed to either reweight full trajectory distributions \cite{shelton2001policy, jie2010connection, zhao2013efficient, metelli2018policy} or to reweight individual state-action pairs \cite{munos2016safe, espeholt2018impala}.
Except for \cite{jie2010connection}, no global IS estimator is derived, but estimates are only based on the current iteration's data.
In contrast, DD-OPG introduces global surrogate model based on all available deterministic policy rollouts and computes local, stochastic approximations using prioritized replay.
Instead of DD-OPG's action space lengthscale, alternative appraoches consider truncation of the importance weights \cite{wawrzynski2007truncated, schulman2017proximal, espeholt2018impala}.
So far, the connection between both approaches has not yet been subject of greater analysis.

Concepts for policy updates range from standard gradient ascent \cite{williams1992simple}, to trust region methods \cite{schulman2015trust} to lower bounds, which can be fully optimized till convergence \cite{schulman2017proximal, metelli2018policy}.
The proposed DD-OPG optimizes a stochastic version based on the lower bound, derived in \cite{metelli2018policy}.

Deterministic policies as means of variance reduction have been previously discussed for example in \cite{sehnke2008policy, plappert2017parameter}.
Instead of action noise for exploration, exploration is achieved by stochasticity in parameter space
The DD-OPG method relies on deterministic policies for variance reduction, but introduces exploration by means of stochastic gradients from the prioritized replay model.

\section{Discussion}
\label{sec:Discussion}

This work presents a new surrogate model of the RL return distribution inspired by importance sampling. 
It can incorporate off-policy data and \emph{deterministic} rollouts to reduce estimator variance.
Despite the promising results and the data-efficient learning progress, several interesting topics remain for future work.

The proposed surrogate model is motivated by its close connections to the importance sampling estimator, the interpretability of the model assumption in action space and its desirable behaviour in the model limits.
A detailed analysis of the resulting model assumptions in policy space, implied by the model assumptions in action space and an analysis of the resulting bias remains an open question.

The proposed optimization scheme empirically achieved good performance in our benchmark experiments, outperforming state-of-the-art methods, although no additional parametric value function baseline (as in TRPO/PPO) is employed. 
However, extensions to other strategies for exploration vs. exploitation, for example acquisition functions like Expected Improvement or Probability of Improvement from Bayesian Optimization \cite{snoek2012practical}, are to be explored and directly carry over to the proposed surrogate return model.

Finally, memory selection is required to scale the non-parametric model structure to typical deep RL applications.
The proposed prioritized trajectory replay is only one possible option to address this challenge.

\newpage
\bibliography{references}
\bibliographystyle{icml2019}

\clearpage

\appendix

\section{Proof of Proposition 1}
\label{sec:ProofofTheorem1}

The weighted importance sampling estimator of the expected cost is given by
\begin{equation}
\hat{J}^{\text{WIS}}(\theta) = \frac{1}{\sum_{i=0}^M w(\tau_i, \theta)} \sum_{i=0}^M w(\tau_i, \theta) R(\tau_i)\,,
\label{eq:WeightedImportanceSampling}
\end{equation}
as derived in Sec.\,3.
Talking the derivative with respect to the policy parameters, we obtain the policy gradient formulation from theorem 1 as shown in \eqref{eq:WeightedImportanceSamplingPolicyGradient}.

\begin{figure*}[t!]
\caption{Derivation of the weighted IS policy gradient.}
\begin{align}
\nabla_\theta \hat{J}^{\text{WIS}}(\theta)
&= 
\nabla_\theta \left( \left( \sum_{i=1}^N \frac{p(\tau_i \mid \theta)}{\frac{1}{N}\sum_j p(\tau_i \mid \theta_j)}\right)^{-1} \right) \sum_{i=0}^N \frac{p(\tau_i \mid \theta)}{\frac{1}{N}\sum_j p(\tau_i \mid \theta_j)} R(\tau_i) + \nonumber \\
& \quad \left( \sum_{i=1}^N \frac{p(\tau_i \mid \theta)}{\frac{1}{N}\sum_j p(\tau_i \mid \theta_j)}\right)^{-1} \sum_{i=0}^N \nabla_\theta \left( \frac{p(\tau_i \mid \theta)}{\frac{1}{N}\sum_j p(\tau_i \mid \theta_j)} R(\tau_i) \right)\\
&= - \left( \sum_{i=1}^N w_i(\theta) \right)^{-2} \left( \sum_{i=1}^N \nabla_\theta w_i(\theta) \right) \left( \sum_{i=1}^N w_i(\theta) R(\tau_i) \right) + \nonumber \\
& \quad \left( \sum_{i=1}^N w_i(\theta) \right)^{-1} \left( \sum_{i=1}^N \nabla_\theta w_i(\theta) R(\tau_i) \right) \\
&= - \frac{1}{Z^2} \left( \sum_{i=1}^N \nabla_\theta w_i(\theta) \right) \left( \sum_{i=1}^N w_i(\theta) R(\tau_i) \right) + \frac{1}{Z} \left( \sum_{i=1}^N \nabla_\theta w_i(\theta) R(\tau_i) \right) \\
&= \frac{1}{Z} \left( \sum_{i=1}^N \nabla_\theta w_i(\theta) R(\tau_i) - \sum_{i=1}^N \nabla_\theta w_i(\theta) \frac{\sum_{i=1}^N w_i(\theta) R(\tau_i)}{Z} \right) \\
&= \frac{1}{Z} \left( \sum_{i=1}^N \nabla_\theta w_i(\theta) R(\tau_i) - \sum_{i=1}^N \nabla_\theta w_i(\theta) \hat{J}^{\text{WIS}}(\theta) \right) \\
\nabla_\theta \hat{J}^{\text{WIS}}(\theta) &= \frac{1}{Z} \sum_{i=1}^N \nabla_\theta w_i(\theta) \left( R(\tau_i) - \hat{J}^{\text{WIS}}(\theta) \right)
\label{eq:WeightedImportanceSamplingPolicyGradient}
\end{align}
\end{figure*}

\section{Experimental Details}
\label{sec:ExperimentalDetails}

In the following section, details about the reference implementations of REINFORCE, TRPO and PPO and their parameter settings are summarized for the benchmark experiments and the ablation study.
Information about the benchmark environments is given in Sec.\,\ref{sec:BenchmarkEnvironments}

\subsection{Algorithm Configurations}
\label{sec:AlgorithmConfigurations}

The reference implementations of the benchmark algorithms REINFORCE, TRPO and PPO are from the Garage RL framework \cite{duan2016benchmarking}. 
A hyper-parameter grid search has been conducted for each algorithm and each environment on separate random seeds.
The parameter ranges and selected hyper-parameters are indicated in Tab.\,\ref{tab:ParameterCartpole}.
For the benchmark itself, ten runs have been conducted for each algorithm and each environment on the random seeds (404, 931, 159, 380, 858, 708, 16, 448, 136, 989).

The configuration of the DD-OPG method is summarized in Tab.\,\ref{tab:ParameterCartpole}.

\begin{table}[t]
\caption{Algorithm hyper-parameters for the benchmark tasks.}
\label{tab:ParameterCartpole}
\vskip 0.15in
\begin{center}
\begin{small}
\begin{sc}
\begin{tabular}{llll}
\toprule
Algorithm & Parameter & Range & Selected \\
\midrule
REINFORCE    & Batch size & [400, 5000] & 5000 \\
             & Step size & [0.0001, 0.1] & 0.03 \\
TRPO         & Batch size & [400, 5000] & 5000 \\
             & Step size & [0.0001, 0.1] & 0.1 \\
PPO          & Batch size & [400, 5000] & 2000 \\
             & Step size & [0.0001, 0.2] & 0.2 \\
\bottomrule
& & & \\
\toprule
Algorithm & Parameter & Symbol & Selected \\
\midrule
DD-OPG & Temperature & $\lambda$ & 0.1 \\
       & Penalty & $\gamma$ & 0.05 \\
       & Lengthscale & $\log \Sigma$ & $3\mathbf{I}$ \\
       & Path buffer & $N_{max}$ & 50 \\
\end{tabular}
\end{sc}
\end{small}
\end{center}
\vskip -0.1in
\end{table}

\begin{table}[t]
\caption{Information about the benchmark environments.}
\label{tab:Environments}
\vskip 0.15in
\begin{center}
\begin{small}
\begin{sc}
\begin{tabular}{llll}
\toprule
Environment & Inputs & States & Horizon \\
\midrule
Cartpole & 1 & 4 & 100 \\
Mountaincar & 1 & 2 & 500 \\
Swimmer & 2 & 13 & 1000 \\
\bottomrule
\end{tabular}
\end{sc}
\end{small}
\end{center}
\vskip -0.1in
\end{table}

\subsection{Benchmark Environments}
\label{sec:BenchmarkEnvironments}

The benchmark environments are cartpole, mountaincar and swimmer from the Garage RL framework.
Details about the input and state dimensions, as well as the task horizons are listed in Tab.\,\ref{tab:Environments}.

\end{document}